\documentclass{elsart}

\usepackage[dvips]{graphicx}
\usepackage{setspace}

\newcommand{\be}{\begin{equation}}
\newcommand{\ee}{\end{equation}}
\newcommand{\beq}{\begin{equation}}
\newcommand{\eeq}{\end{equation}}
\newcommand{\bed}{\begin{displaymath}}
\newcommand{\eed}{\end{displaymath}}
\newcommand{\beqa}{\begin{eqnarray}}
\newcommand{\eeqa}{\end{eqnarray}}
\newcommand{\beqann}{\begin{eqnarray*}}
\newcommand{\eeqann}{\end{eqnarray*}}
\newcommand{\bseq}{\begin{subequations}}
\newcommand{\eseq}{\end{subequations}}

\newcommand{\ba}{\begin{array}}
\newcommand{\ea}{\end{array}}

\newcommand{\set}[3]{\{\,#1\,\}_{#2}^{#3}}

\newcommand{\tr}{{\rm tr}}

\newcommand{\negr}[1]{{\bf {#1}}}
\newcommand{\nigr}[2]{{\bf {#1}}_{#2}}

\usepackage{subeqn}
 
 \newtheorem{Def}{Definition}
 
 \newtheorem{Lem}{Lemma}

\usepackage{amssymb}

\begin{document}

\begin{frontmatter}

 \title{On the Kinetostatic Optimization of Revolute-Coupled Planar Manipulators}
 \author{Chablat D.}
 \ead{Damien.Chablat@irccyn.ec-nantes.fr}
 \ead[url]{www.irccyn.ec-nantes.fr}
 \thanks[label1]{IRCCyN: UMR 6597 CNRS, Ecole Centrale de Nantes, Universite de Nantes, Ecole des Mines de Nantes}
 \address{Institut de Recherche en Communications et Cybern\'etique de Nantes \thanksref{label1} }
 \address{1, rue de la No\"e, 44321 Nantes, France }

 \author{Angeles J.}
 \address{Department of Mechanical Engineering, \goodbreak
   Centre for Intelligent Machines, McGill University, \goodbreak
   817 Sherbrooke Street West, Montreal, Quebec, Canada H3A 2K6 }
 \ead{angeles@cim.mcgill.ca}
 \ead[url]{www.cim.mcgill.ca}

\begin{abstract}
Proposed in this paper is a kinetostatic performance index for the
optimum dimensioning of planar manipulators of the serial type.
The index is based on the concept of {\em distance} of the
underlying Jacobian matrix to a given isotropic matrix that is
used as a reference model for purposes of performance evaluation.
Applications of the index fall in the realm of design, but control
applications are outlined. The paper focuses on planar
manipulators, the basic concepts being currently extended to their
three-dimensional counterparts.\end{abstract}

\end{frontmatter}

\section{Introduction}
Various performance indices have been devised to asses the
kinetostatic performance of serial manipulators. Among these, the
concepts of {\em service angle} \cite{Vinogradov:71}, {\em
dexterous workspace} \cite{Kumar:81} and {\em manipulability}
\cite{Yoshikawa:85} are worth mentioning. All these different
concepts allow the definition of the kinetostatic performance of a
manipulator from correspondingly different viewpoints. However,
with the exception of Yoshikawa's manipulability index
\cite{Yoshikawa:85}, none of these considers the invertibility of
the Jacobian matrix. A dimensionless quality index was recently
introduced by Lee \cite{Lee:98} based on the ratio of the Jacobian
determinant to its maximum absolute value, as applicable to
parallel manipulators. This index does not take into account the
location of the operation point in the end-effector, for the
Jacobian determinant is independent of this location. The proof of
the foregoing fact is available in \cite{Angeles:97}, as
pertaining to serial manipulators, its extension to their parallel
counterparts being straightforward. The {\em condition number} of
a given matrix is well known to provide a measure of invertibility
of the matrix \cite{Golub:89}. It is thus natural that this
concept found its way in this context. Indeed, the condition
number of the Jacobian matrix was proposed by Salisbury
\cite{Salisbury:82} as a figure of merit to minimize when
designing manipulators for maximum accuracy. In fact, the
condition number gives, for a square matrix, a measure of the
relative roundoff-error amplification of the computed results
\cite{Golub:89} with respect to the data roundoff error. As is
well known, however, the dimensional inhomogeneity of the entries
of the Jacobian matrix prevents the straightforward application of
the condition number as a measure of Jacobian invertibility. The
{\em characteristic length} was introduced in \cite{Angeles:92} to
cope with the above-mentioned inhomogeneity. Apparently,
nevertheless, this concept has found strong opposition within some
circles, mainly because of the lack of a direct geometric
interpretation of the concept. It is the aim of this paper to shed
more light in this debate, while proposing a novel performance
index that lends itself to a straightforward manipulation and
leads to sound geometric relations. Briefly stated, the
performance index proposed here is based on the concept of {\em
distance} in the space of $m\times n$ matrices, which is based, in
turn, on the concept of inner product of this space. The
performance index underlying this paper thus measures the distance
of a given Jacobian matrix from an isotropic matrix of the same
gestalt. With the purpose of rendering the Jacobian matrix
dimensionally homogeneous, we resort to the concept of
posture-dependent {\em conditioning length}. Thus, given an
arbitrary serial manipulator in an arbitrary posture, it is
possible to define a unique length that renders this matrix
dimensionally homogeneous and of minimum distance to isotropy. The
characteristic length of the manipulator is then defined as the
conditioning length corresponding to the posture that renders the
above-mentioned distance a minimum over all possible manipulator
postures. This paper is devoted to planar manipulators, the
concepts being currently extended to spatial ones.
\section{Algebraic Background}
Given two arbitrary $m\times n$ matrices \negr A and \negr B of
real entries, their inner product, represented by $(\negr
A,\,\negr B)$, is defined as
 \beq
  (\negr A,\,\negr B)\equiv {\rm tr}(\negr A\negr W\negr B^T)
  \label{e:inner-product}
 \eeq
where \negr W is a {\em positive-definite} $n\times n$ weighting
matrix that is introduced to allow for suitable normalization. The
entries of \negr W need not be dimensionally homogeneous, and, in
fact, they should not if \negr A and \negr B are not. However, the
product $\negr A \negr W \negr B^T$ must be dimensionally
homogeneous; else, its trace is meaningless. The {\em norm} of the
space of $m\times n$ matrices induced by the above inner product
is thus the {\em Frobenius norm}, namely,
 \bseq
 \beq
  \|\negr A\|^2 = {\rm tr}(\negr A\negr W\negr A^T)
  \label{e:norm}
 \eeq
Moreover, we shall be handling only nondimensional matrix entries,
and hence, we choose \negr W non-dimensional as well, and so as to
yield a value of unity for the norm of the $n\times n$ identity
matrix \negr 1. Hence,
 \beq
  \negr W\equiv \frac{1}{n}\negr 1
  \label{e:weight}
 \eeq
The foregoing inner product is thus expressed as
 \beq
  (\negr A,\,\negr B)\equiv
  \frac{1}{n}{\rm tr}(\negr A\negr B^T)
  \label{e:W-inner-product}
 \eeq
 \eseq
Henceforth we shall use only the Frobenius norm; for brevity, this
will be simply referred to as the {\em norm} of a given matrix.
\par
When comparing two dimensionless $m\times n$ matrices \negr A and
\negr B, we can define their distance $d(\negr A,\,\negr B)$ as
the Frobenius norm of their difference, namely,
 \bseq
  \beq
  d(\negr A,\,\negr B)\equiv \|\negr A - \negr B\|
  \label{e:distance}
  \eeq
i.e.,
 \beq
  d(\negr A,\,\negr B)\equiv \sqrt{\frac{1}{n}{\rm tr}
  [(\negr A - \negr B) (\negr A - \negr B)^T]}
  \label{e:distance-explicit}
  \eeq
 \eseq
\par
An $m\times n$ isotropic matrix, with $m<n$, is one with a
singular value $\sigma>0$ of multiplicity $m$, and hence, if the
$m\times n$ matrix \negr C is isotropic, then
 \beq
  \negr C\negr C^T =
  \sigma^2\negr 1\label{e:isotropic}
 \eeq
where \negr 1 is the $m\times m$ identity matrix. Note that the
generalized inverse of \negr C can be computed without
roundoff-error, for it is proportional to $\negr C^T$, namely,
\beq (\negr C\negr C^T)^{-1}\negr C^T=\frac{1}{\sigma^2}\negr C^T
\label{e:gen-inverse} \eeq
\par
\begin{figure}[hbt]
\begin{center}
\includegraphics[width=60mm,height=50mm]{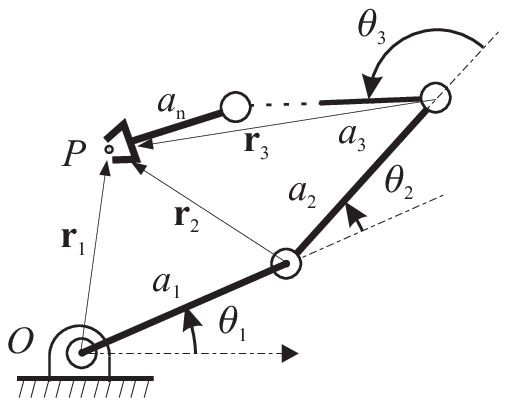}
\caption{Planar $n$-revolute manipulator}
\protect\label{figure:planar_n_R_manipulator}
\end{center}
\end{figure}

Furthermore, the condition number $\kappa(\negr A)$ of a square
matrix \negr A is defined \cite{Golub:89} as
 \beqa
   \kappa(\negr A)=
   \|\negr A\| \|\negr A^{-1}\| \label{equation:kappa}
 \eeqa
where {\em any} norm can be used. For purposes of the paper, we
shall use the Frobenius norm for matrices and the Euclidian norm
for vectors. Henceforth we assume, moreover, a planar $n$-revolute
manipulator, as depicted in
Fig.~\ref{figure:planar_n_R_manipulator}, with Jacobian matrix
\negr J given by Angeles \cite{Angeles:97}
 \beqa \negr J=
 \left[\begin{array}{cccc} 1 & 1 & \cdots & 1 \\ \negr E \negr r_1
 & \negr E \negr r_2 & \cdots & \negr E \negr r_n
 \end{array}
 \right]
 \label{equation:jacobian_matrix}
 \eeqa
where $\negr r_i$ is the vector directed from the center of the
$i$th revolute to the operation point $P$ of the end-effector,
while matrix $\negr E$ is defined as
 \beqa
  \negr E= \left[\begin{array}{cc}
  0 & -1 \\
  1 & 0
  \end{array}
  \right]
  \label{e:E}
\eeqa i.e., \negr E represents a counterclockwise rotation of
$90^\circ$. It will prove convenient to partition {\negr J} in the
form \beqa \negr J = \left[\begin{array}{c} \negr A \\ \negr B
\end{array}
\right] \nonumber \eeqa with \negr A and \negr B defined as \beqa
\negr A = \left[\begin{array}{cccc} 1 & 1 & \cdots & 1
\end{array}
\right] {\rm ~~~~and~~~~} \negr B = \left[\begin{array}{cccc}
\negr E \negr r_1 & \negr E \negr r_2 & \cdots & \negr E \negr r_n
\end{array}
\right] \nonumber \eeqa

Therefore, while the entries of \negr A are dimensionless, those
of \negr B have units of length. Thus, the sole singular value of
\negr A, i.e. the nonnegative square root of the scalar of $\negr
A \negr A^T$, is $\sqrt{n}$, and hence, dimensionless, and
pertains to the mapping from joint-rates into end-effector angular
velocity. The singular values of \negr B, which are the
nonnegative square roots of the eigenvalues of $\negr B \negr
B^T$, have units of length, and account for the mapping from
joint-rates into operation-point velocity. It is thus apparent
that the singular values of $\negr J$ have different dimensions
and hence, it is impossible to compute $\kappa(\negr J)$ as in
eq.(\ref{equation:kappa}), for the norm of $\negr J$, as defined
in eqs.(\ref{e:norm} \& b), is meaningless. The normalization of
the Jacobian for purposes of rendering it dimensionless has been
judged to be dependent on the normalizing length \cite{Paden:88}.
As a means to avoid the arbitrariness of the choice of that
normalizing length, the characteristic length $L$ was introduced
in \cite{Ranjbaran:95}. Since the calculation of $L$ is based on
the minimization of an objective function that is elusive to a
straightforward geometric interpretation, namely, the condition
number of the normalized Jacobian, the characteristic length has
been found cumbersome to use in manipulator design. We introduce
below the concept of {\em conditioning length} to render the
Jacobian matrix dimensionless, which will allow us to define the
characteristic length using a geometric approach. In the sequel,
we will need the partial derivative of the trace of a square
matrix \negr N with respect to a scalar argument $x$ of \negr N.
The said derivative is readily obtained as \bseq \beq
\frac{\partial}{\partial x}\tr(\negr N)=\tr\left(\frac{\partial
\negr N}{\partial x}\right)\label{e:partial-trace} \eeq Moreover,
in some instances, we will need the partial derivative of a scalar
function $f$ of the matrix argument \negr N with respect to the
scalar $x$, which is, in turn, an argument of \negr N. In this
case, the desired partial derivative is obtained by application of
the {\em chain rule}: \beq \frac{\partial f}{\partial x} =
\tr\left(\frac{\partial f} {\partial\negr N}\frac{\partial\negr
N^T}{\partial x}\right) \label{e:partialf/partialx} \eeq In
particular, when $f(\negr N)$ is the $k$th {\em moment} $N_k$ of
\negr N with respect to $x$, defined as \beq N_k\equiv\tr(\negr
N^k) \label{e:kth-moment} \eeq the partial derivative of $N_k$
with respect to $x$ is given by \beq \frac{\partial N_k}{\partial
x}=k\tr\left(\negr N^{k-1}\frac{\partial \negr N^T}{\partial
x}\right)\label{e:partial-kth-moment} \eeq \eseq Furthermore, we
recall that the trace of any square matrix \negr N equals that of
its transpose, i.e., \bseq \beq \tr(\negr N^T) =\tr(\negr
N)\label{e:trace-transp} \eeq and, finally, the trace of a product
of various matrices compatible under multiplication does not
change under a cyclic permutation of the factors, i.e., if \negr
A, \negr B, and \negr C are three matrices whose product $\negr
A\negr B\negr C$ is possible and square, then \tr(\negr A\negr
B\negr C) =\tr(\negr B\negr C\negr A)= \tr(\negr C\negr A\negr B)
\label{e:product-trace} \eseq
\section{Isotropic Sets of Points}
Consider the set ${\mathcal S}\equiv \set{P_k}1n$ of $n$ points in
the plane, of position vectors $\set{\negr p_k}1n$, and centroid
$C$, of position vector \negr c, i.e., \beq \negr
c\equiv\frac{1}{n}\sum_1^n \negr p_k \label{e:centroid} \eeq The
summation appearing in the right-hand side of the above expression
is known as the {\em first moment of $\mathcal S$ with respect to
the origin $O$} from which the position vectors stem. The {\em
second moment of $\mathcal S$ with respect to $C$} is defined as a
tensor \negr M, namely, \beq \negr M\equiv \sum_1^n (\nigr
pk-\negr c)(\nigr pk-\negr c)^T \label{e:second} \eeq It is now
apparent that the {\em root-mean square} value of the distances
$\set{d_k}1n$ of $\mathcal S$, $d_{\rm rms}$, to the centroid is
directly related to the trace of \negr M, namely, \beq d_{\rm
rms}\equiv\sqrt{\frac{1}{n}\sum_1^n (\nigr pk-\negr c)^T (\nigr
pk-\negr c)} \equiv \sqrt{\frac{1}{n}\tr(\negr M)} \label{e:d_rms}
\eeq Further, the {\em moment of inertia} \negr I of $\mathcal S$
with respect to the centroid is defined as that of a set of unit
masses located at the points of $\mathcal S$, i.e.,
 \bseq
  \beq
  \negr I\equiv \sum_1^n [\,\|\nigr pk-\negr c\|^2\negr 1 - (\nigr pk
  -\negr c)(\nigr pk-\negr c)^T\,]\label{e:inertia}
 \eeq
in which \negr 1 is the $2\times 2$ identity matrix. Hence, in
light of definitions (\ref{e:second}) and (\ref{e:d_rms}),
 \beq
  \negr I= \tr(\negr M) \negr 1 - \negr M\label{e:inertia-alt}
  \eeq
 \eseq
We shall refer to {\bf I} as the {\em geometric moment of inertia}
of $\mathcal S$ about its centroid. It is now apparent that \negr
I is composed of two parts, an isotropic matrix of norm $\tr(\negr
M)$ and the second moment of $\mathcal S$ with the sign reversed.
Moreover, the moment of inertia \negr I can be expressed in a form
that is more explicitly dependent upon the set $\set{\nigr pk-
\negr c}1n$, if we recall the concept of {\em cross-product
matrix} \cite{Angeles:97}. Briefly stated, for any
three-dimensional vector \negr v, we define the cross-product
matrix $\nigr Pk$ of $(\nigr pk - \negr c)$, or of any other
three-dimensional vector for that matter, as
 \bseq
  \beq
  \nigr Pk\equiv \frac{\partial [(\nigr pk-\negr c)\times\negr v]}
  {\partial\negr v}
  \label{e:CPM}
 \eeq
Further, we recall the identity \cite{Angeles:97}
 \beq
  \negr P_k^2\equiv -\|\nigr pk-\negr c\|^2\negr 1 + (\nigr pk-\negr c)
  (\nigr pk-\negr c)^T
  \label{e:CPM^2}
  \eeq
 \eseq
\par\noindent It is now apparent that the moment of inertia of $\mathcal
S$ takes the simple form
 \beq
  \negr I = - \sum_1^n \negr P_k^2
  \label{e:inertia-CPM}
 \eeq
We thus have
\begin{Def}[Isotropic Set]
The set $\mathcal S$ is said to be isotropic if its second-moment
tensor with respect to its centroid is isotropic.
\end{Def}
As a consequence, we have
\begin{Lem}
The geometric moment of inertia of an isotropic set of points
about its centroid is isotropic.
\end{Lem}
\subsection{Geometric Properties of Isotropic Sets of Points}
We describe below some properties of isotropic sets of points that
will help us better visualize the results that follow.
\subsubsection{Union of Two Isotropic Sets of Points}
Consider two isotropic sets of points in the plane, ${\mathcal
S}_1= \set{P_k}1n$ and ${\mathcal S}_2=\set{P_k}{n+1}{n+m}$. If
the centroid $C$ of the position vector $\negr c$ of ${\mathcal
S}_1$ coincides with that of ${\mathcal S}_2$, i.e. if, \beqa
\negr c \equiv \frac{1}{n} \sum_1^n \negr p_k \equiv
\frac{1}{m}\sum_{n+1}^{n+m} \negr p_k \eeqa then, the set
${\mathcal S}={\mathcal S}_1 \cup {\mathcal S}_2$ is isotropic.
\par
\begin{figure}[hbt]
\begin{center}
\includegraphics[width=140mm,height=40mm]{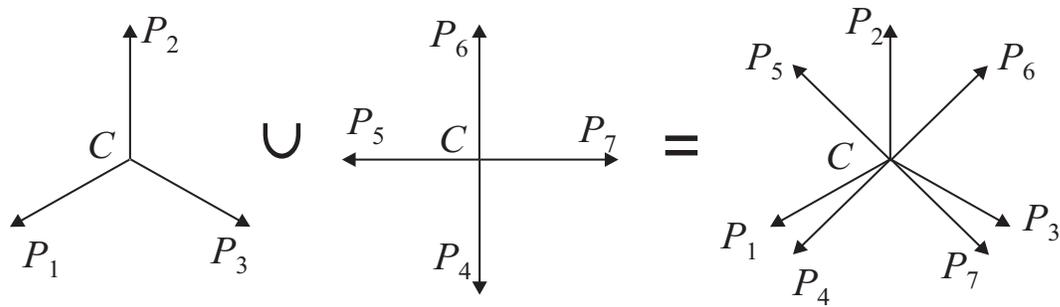}
\caption{Union of two isotropic sets of points}
\protect\label{figure:addition_points}
\end{center}
\end{figure}

For example, let ${\mathcal S}_1$ be a set of three isotropic
points and ${\mathcal S}_2$ a set of four isotropic points, as
displayed in Fig.~\ref{figure:addition_points}, i.e.,
 \bseq
 \beqa
  {\mathcal S}_1&=&\left\{
  \left[\begin{array}{c}
  - \sqrt{6} /2 \\
  - \sqrt{2} /2
  \end{array}
  \right],
  \left[\begin{array}{c}
  ~ \sqrt{6} /2 \\
  - \sqrt{2} /2
  \end{array}
  \right],
  \left[\begin{array}{c}
  0 \\ \sqrt{2}
  \end{array}
  \right]
  \right\} \nonumber \\
  {\mathcal S}_2&=& \left\{
  \left[\begin{array}{c}
  0 \\ - \sqrt{2}
  \end{array}
  \right],
  \left[\begin{array}{c}
  - \sqrt{2} \\ 0
  \end{array}
  \right],
  \left[\begin{array}{c}
  0 \\ \sqrt{2}
  \end{array}
  \right],
  \left[\begin{array}{c}
  \sqrt{2} \\ 0
  \end{array}
  \right]
  \right\}
  \eeqa
where the centroid $C$ is the origin. The second moment of
$\mathcal S$ with respect to $\mathcal C$ is isotropic, namely,
 \beq
  \negr M \equiv \sum_1^{3+4} (\nigr pk-\negr c)(\nigr pk-\negr c)^T
  =\left[\begin{array}{cc}
   7 & 0 \\
   0 & 7
  \end{array}
  \right]=
  (7)\negr 1
 \eeq
 \eseq
where \negr 1 denotes the $2\times 2$ identity matrix.
Furthermore, the geometric moment of inertia of $\mathcal S$ is
 \beqa
   \negr 1=(14)\negr 1 - (7)\negr 1 = (7)\negr 1
   \nonumber
 \eeqa

\begin{Lem}
The union of two isotropic sets of points sharing the same
centroid is also isotropic.
\end{Lem}
\subsubsection{Rotation of an Isotropic Set of Points}
\begin{figure}[hbt]
\begin{center}
\includegraphics[width=130mm,height=40mm]{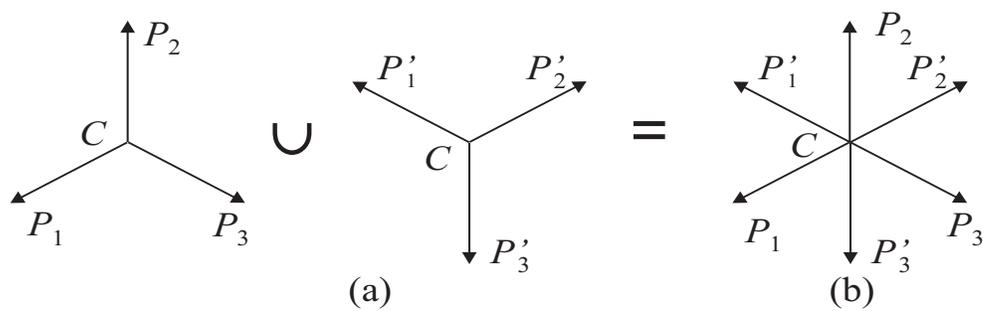}
\caption{(a) Rotation of an isotropic set $\mathcal S$ and (b)
union of $\mathcal S$ with its rotated image}
\protect\label{figure:rotation_points}
\end{center}
\end{figure}

Let $\negr R$ denote a rotation matrix in the plane through an
angle $\alpha$ and ${\mathcal S}=\set{P_k}1n$ a set of isotropic
points. A new set of points ${\mathcal S'}=\set{P'_k}1n$ is
defined upon rotating ${\mathcal S}$ through an angle $\alpha$
about $C$ as a rigid body. The second moment of $\mathcal S'$ with
respect to $C$ is shown below to be isotropic as well. Indeed,
letting this moment be $\negr M^\prime$, we have
 \beqa
  \negr M^\prime \equiv \sum_1^n (\negr p_k^\prime - \negr c^\prime)
  (\negr p_k^\prime - \negr c^\prime)^T
  \nonumber
 \eeqa
where, by definition,
 \beqa
  \negr p_k^\prime - \negr c^\prime=\negr R(\negr p_k-\negr c)
  \nonumber
 \eeqa
Thus,
 \beqa
  \negr M^\prime \equiv \sum_1^n \negr R(\negr p_k - \negr c)(\negr p_k -
  \negr c)^T \negr R^T = \negr R[\,\sum_1^n (\negr p_k - \negr c)(\negr
  p_k - \negr c)^T] \negr R^T
  \nonumber
 \eeqa
But the summation in brackets is the second moment \negr M of the
set $\mathcal S$, which is, by assumption, isotropic, and hence,
takes the form
 \beqa
  \negr M = \sum_1^n (\negr p_k - \negr c)(\negr p_k - \negr c)^T=
  \sigma^2\negr 1
  \nonumber
 \eeqa
for a real number $\sigma > 0$ and \negr 1 denoting the $2\times
2$ identity matrix. Hence,
 \beqa
  \negr M^\prime = \sigma^2\negr R\negr R^T =\sigma^2 \negr 1
  \nonumber
 \eeqa
thereby proving that the rotated set is isotropic as well. We thus
have
\begin{Lem}
The rotation of an isotropic set of points as a rigid body with
respect to its centroid is also isotropic.
\end{Lem}
The counterclockwise rotation of an isotropic set of three points,
$\mathcal S$, through an angle of $60^\circ$ and the union of the
original set and its rotated counterpart are depicted in
Fig.~\ref{figure:rotation_points}. Note that the union of the two
sets is isotropic as well.
\subsubsection{Trivial Isotropic Set of Points}
\begin{figure}[hbt]
\begin{center}
\includegraphics[width=90mm,height=80mm]{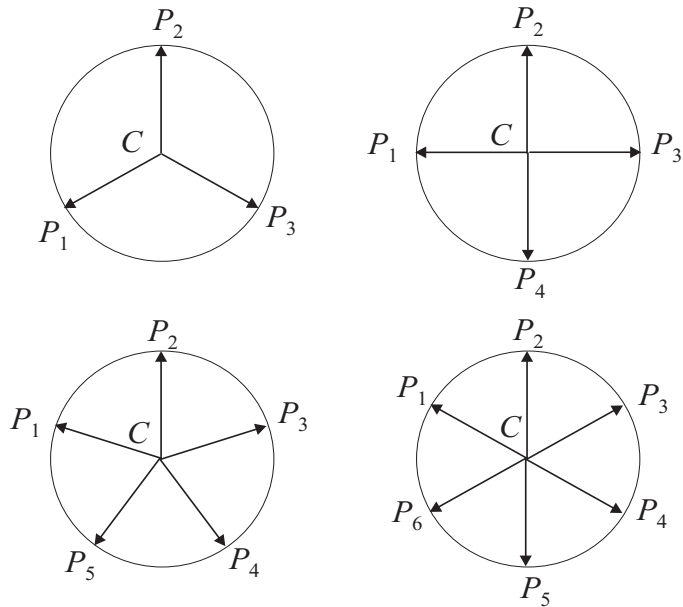}
\caption{Trivial isotropic sets for $n=3, \cdots, 6$}
\protect\label{figure:Trivial}
\end{center}
\end{figure}

An isotropic set of points can be defined by the union or
rotation, or a combination of both, of isotropic sets. The
simplest set of isotropic points is the set of vertices of a
regular polygon. We thus have
\begin{Def}[Trivial isotropic set]
A set of $n$ points ${\mathcal S}$ is called {\em trivial} if it
is the set of vertices of a regular polygon with $n$ vertices.
\end{Def}
Trivial isotropic sets, ${\mathcal S}= \set{P_k}1n$ are depicted
in Fig.~\ref{figure:Trivial}, for $n=3, \cdots, 6$.

Also note that
\begin{Lem}
A trivial isotropic set $\mathcal S$ remains isotropic under every
reflection about an axis passing through the centroid $C$.
\end{Lem}
\section{An Outline of Kinematic Chains}
The connection between sets of points and planar manipulators of
the serial type is the concept of {\em simple kinematic chain}.
For completeness, we recall here some basic definitions pertaining
to this concept.
\subsection{Simple Kinematic Chains}
The kinematics of manipulators is based on the concept of {\em
kinematic chain}. A kinematic chain is a set of {\em rigid
bodies}, also called {\em links}, coupled by {\em kinematic
pairs}. In the case of planar chains, two lower kinematic pairs
are possible, the revolute, allowing pure rotation of the two
coupled links, and the prismatic pair, allowing a pure relative
translation, along one direction, of the same links. For the
purpose of this paper, we study only revolute pairs, but prismatic
pairs are also common in manipulators.
\par
\begin{Def}[Simple kinematic chain]
A kinematic chain is said to be {\em simple} if each and every one
of its links is coupled to at most two other links.
\end{Def}
A simple kinematic chain can be {\em open} or {\em closed\/}; in
studying serial manipulators we are interested in the former. In
such a chain, we distinguish exactly two links, the terminal ones,
coupled to only one other link. These links are thus said to be
{\em simple}, all other links being {\em binary}. In the context
of manipulator kinematics, one terminal link is arbitrarily
designated as {\em fixed}, the other terminal link being the {\em
end-effector} (EE), which is the one executing the task at hand.
The task is defined, in turn, as a sequence of {\em
poses}---positions and orientations---of the EE, the position
being given at a specific point $P$ of the EE that we term the
{\em operation point}.
\subsection{Isotropic Kinematic Chains}
To every set $\mathcal S$ of $n$ points it is possible to
associate a number of kinematic chains. To do this, we number the
points from 1 to $n$, thereby defining $n-1$ links, the $i$th link
carrying joints $i$ and $i+1$. Links are thus correspondingly
numbered from 1 to $n$, the $n$th link, or EE, carrying joint $n$
on its proximal (to the base) end and the operation point $P$ on
its distal end. Furthermore, we define an additional link, the
base, which is numbered as 0.
\par
It is now apparent that, since we can number a given set $\mathcal
S$ of $n$ points in $n!$ possible ways, we can associate $n!$
kinematic chains to the above set $\mathcal S$ of $n$ points.
Clearly, these chains are, in general, different, for the lengths
of their links are different as well. Nevertheless, some pairs of
identical chains in the foregoing set are possible.
\begin{Def}[Isotropic kinematic chain]
If the foregoing set $\mathcal S$ of $n$ points is isotropic, and
the operation point $P$ is defined as the centroid of $\mathcal
S$, then any kinematic chain stemming from $\mathcal S$ is
isotropic.
\end{Def}
\section{The Posture-Dependent Conditioning Length of
Planar n-Revolute Manipulators}
Under the assumption that the manipulator finds itself at a {\em
posture} $\mathcal P$ that is given by its set of joint angles,
$\set{\theta_k}1 n$, we start by dividing the last $n$ rows of the
Jacobian by a length $l_{\mathcal P}$, as yet to be determined.
This length will be found so as to minimize the distance of the
normalized Jacobian to a corresponding isotropic matrix \negr K,
subscript $ \mathcal P$ reminding us that, as the manipulator
changes its posture, so does the length $l_{\mathcal P}$. This
length will be termed the {\em conditioning length} of the
manipulator at $\mathcal P$.
\subsection{A Dimensionally-Homogeneous Jacobian Matrix}
In order to distinguish the original Jacobian matrix from its
dimensionally-homogeneous counterpart, we shall denote the latter
by $\overline{\negr J}$, i.e.,
 \beqa
  \overline{\negr J}=
  \left[\begin{array}{cccc}
  1 & 1 & \cdots & 1 \\
  (1/ l_{\mathcal P})~ \negr E \negr r_1 &
  (1/ l_{\mathcal P})~ \negr E \negr r_2 &
  \cdots &
  (1/ l_{\mathcal P})~ \negr E \negr r_n
  \end{array}
  \right] \nonumber
 \eeqa
Now the conditioning length will be defined via the minimization
of the distance of the dimensionally-homogeneous Jacobian matrix
$\overline{\negr J}$ of an $n$-revolute manipulator to an
isotropic $3\times n$ {\em model matrix} \negr K whose entries are
dimensionless and has the same gestalt as any $3\times n$ Jacobian
matrix. To this end, we define an isotropic set ${\mathcal
K}=\set{K_i}1n$ of $n$ points in a {\em dimensionless} plane, of
position vectors $\set{\nigr ki}1n$, which thus yields the
dimensionless matrix
 \beq
  \negr K=
  \left[\begin{array}{cccc}
  1 & 1 & \cdots & 1 \\
  \negr E\nigr k1 & \negr E\nigr k2 & \cdots & \negr E\negr k_n
  \end{array}
  \right]
  \label{e:K}
 \eeq
Further, we compute the product $\negr K\negr K^T$:
 \beqa
  \negr K \negr K^T=
  \left[\begin{array}{cc}
  n & \sum_1^n \negr k_i^T\negr E^T \\
  \sum_1^n \negr E\negr k_i &
  \sum_1^n \negr E\negr k_i\negr k_i^T\negr E^T
  \end{array}
  \right]
  \nonumber
 \eeqa
Upon expansion of the summations occurring in the above matrix, we
have
 \bseq
 \beqa
  \sum_1^n \negr k_i^T\negr E^T &=& (\sum_1^n \negr E\negr k_i)^T =
  \negr E(\sum_1^n \nigr ki)^T\label{e:sum-k_i}\\
  \sum_1^n \negr E\negr k_i\negr k_i^T\negr E^T&=&
  \negr E(\sum_1^n \negr k_i\negr k_i^T)\negr E^T\label{e:sum-k_ik_i^T}
  \eeqa
 \eseq
Now, by virtue of the assumed isotropy of $\mathcal K$, the terms
in parentheses in the foregoing expressions become
 \beqa
  \sum_1^n \nigr ki&=&\negr 0\nonumber\\
  \sum_1^n \nigr ki\negr k_i^T&=&k^2\nigr 1{2\times 2}\nonumber
 \eeqa
where the factor $k^2$ is as yet to be determined and $\nigr
1{2\times 2}$ denotes the $2\times 2$ identity matrix. Hence, the
product $\negr K\negr K^T$ takes the form
 \beq
  \negr K\negr K^T=
    \left[\begin{array}{cc}
      n         & \negr 0^T \\
      \negr 0   & k^2 \nigr 1{2\times 2}
    \end{array}
    \right]
  \label{e:KK^T}
 \eeq
Now, in order to determine $k^2$, we recall that matrix \negr K is
isotropic, and hence that the product $\negr K\negr K^T$ has a
triple eigenvalue. It is now apparent that the triple eigenvalue
of the said product must be $n$, which means that
 \beq
  k^2=n\label{e:k^2}
 \eeq
and hence,
 \beqa
  \sum_1^n \nigr ki\negr k_i^T=(n)\nigr 1{2\times 2}
  \nonumber
 \eeqa
\subsection{Example~1: A Three-DOF Planar Manipulator}
\label{ss:example-1}
\begin{figure}[hbt]
\begin{center}
\includegraphics[width=135mm,height=44mm]{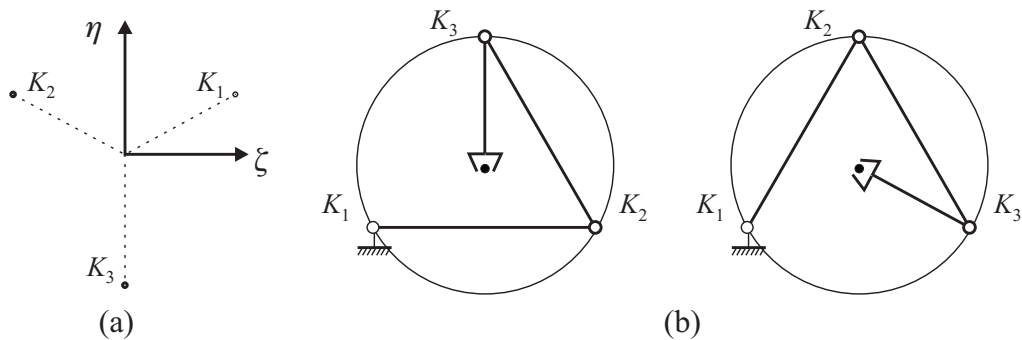}
\caption{Two isotropic manipulator postures stemming from the same
isotropic set upon a relabelling of its points}
\protect\label{figure:3dof_posture_isotrope}
\end{center}
\end{figure}

Shown in Fig.~\ref{figure:3dof_posture_isotrope}a  is an isotropic
set $\mathcal K$ of three points, of position vectors $\set{\negr
k_i}1n$, in a nondimensional plane. The position vectors are given
by \bseq \beq \negr k_1 = \frac{1}{2} \left[\begin{array}{c}
\sqrt{6} \\ \sqrt{2}
\end{array}
\right] {~\rm ,~} \negr k_2 = \frac{1}{2} \left[\begin{array}{c} -
\sqrt{6} \\ ~ \sqrt{2}
\end{array}
\right] {~\rm ,~} \negr k_3 = \left[\begin{array}{c} 0 \\ -
\sqrt{2}
\end{array}
\right] \eeq Hence, the corresponding model matrix is \beq \negr
K= \left[\begin{array}{ccc} 1 & 1 & 1  \\ -\sqrt{2} / 2 &
-\sqrt{2} / 2 & ~\sqrt{2} \\
 ~\sqrt{6} / 2 &
-\sqrt{6} / 2 & 0
\end{array}
\right] \eeq \eseq \noindent which can readily be proven to be
isotropic, with a triple singular value of $k=\sqrt{3}$. When the
order of the three vectors is changed, the isotropic condition is
obviously preserved, for such a reordering amounts to nothing but
a relabelling of the points of $\mathcal K$. Also note that two
isotropic matrices $\negr K$ are associated with two symmetric
postures, as displayed in Fig.~\ref{figure:3dof_posture_isotrope}b
\subsection{Example~2: A Four-DOF Redundant Planar Manipulator}
An isotropic set $\mathcal K$ of four points, $\set{K_i}14$, is
defined in a nondimensional plane, with position vectors $\negr
k_i$ given below: \bseq \beq \negr k_1 = \left[\begin{array}{c}
\sqrt{2} \\ \sqrt{2}
\end{array}
\right] {~\rm ,~} \negr k_2 = \left[\begin{array}{c} - \sqrt{2} \\
~ \sqrt{2}
\end{array}
\right] {~\rm ,~} \negr k_3 = \left[\begin{array}{c} - \sqrt{2} \\
- \sqrt{2}
\end{array}
\right] {~\rm ,~} \negr k_4 = \left[\begin{array}{c} ~ \sqrt{2} \\
- \sqrt{2}
\end{array}
\right] \eeq which thus lead to \beq \negr K=
\left[\begin{array}{cccc} 1 & 1 & 1 & 1\\ -\sqrt{2} & -\sqrt{2} &
\sqrt{2} & \sqrt{2} \\ \sqrt{2} & -\sqrt{2} & -\sqrt{2} & \sqrt{2}
\end{array}
\right] \eeq \eseq

\begin{figure}[hbt]
\begin{center}
\includegraphics[width=85mm,height=110mm]{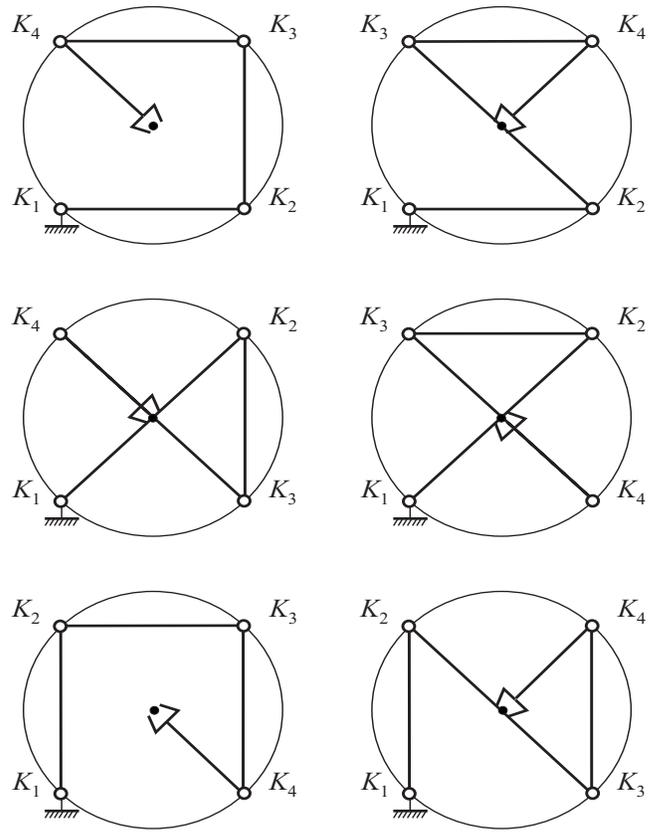}
\caption{Six isotropic postures for the same isotropic set}
\protect\label{figure:4dof_posture_isotrope}
\end{center}
\end{figure}

We thus have $4! = 24$ isotropic kinematic chains for a four-dof
planar manipulator, but we represent only $6$ in
Fig.~\ref{figure:4dof_posture_isotrope} because the choice of the
first point is immaterial, since this choice amounts to a rotation
of the overall manipulator as a rigid body.
\subsection{Computation of the Conditioning Length}
\begin{figure}[hbt]
\begin{center}
\includegraphics[width=70mm,height=35mm]{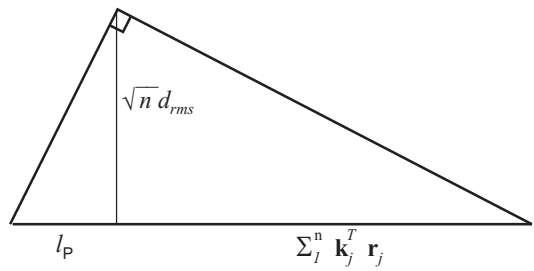}
\caption{A geometric interpretation of the conditioning length}
\protect\label{figure:interpretation}
\end{center}
\end{figure}

We can now formulate a least-square problem aimed at finding the
conditioning length $l_{\mathcal P}$ that renders the distance
from $\overline{\negr J}$ to \negr K a minimum. The task will be
eased if we work rather with the reciprocal of $l_{\mathcal P}$,
$\lambda\equiv 1/l_{\mathcal P}$, and hence,
 \beq
  z\equiv
  \frac{1}{2}
  \frac{1}{n}
  \tr[(\overline{\negr J}-\negr K)
  (\overline{\negr J}-\negr K)^T]
  \quad\rightarrow\quad\min_\lambda
  \label{e:least-squares}
 \eeq
Upon expansion,
 \beqa
  z=\frac{1}{2}
  \frac{1}{n}
  \tr(\overline{\negr J}\overline{\negr J}^T
  -\overline{\negr J}\negr K^T-\negr K\overline{\negr J}^T
  +\negr K\negr K^T)
  \nonumber
 \eeqa
Since the trace of a matrix equals that of its transpose, i.e.,
 \beqa
  \tr(\overline{\negr J}\negr K^T)=\tr(\negr K\overline{\negr J}^T)
  \nonumber
 \eeqa
the foregoing expression for $z$ reduces to \beq z\equiv
\frac{1}{2} \frac{1}{n} \tr(\overline{\negr J}\overline{\negr J}^T
-2\negr K\overline{\negr J}^T+\negr K\negr K^T) \label{e:simple-z}
\eeq It is noteworthy that the above minimization problem is (a)
quadratic in $\lambda$, for $\overline{\negr J}$ is linear in
$\lambda$ and (b) unconstrained, which means that the problem
accepts a unique solution. This solution can be found,
additionally, in closed form. Indeed, the optimum value of
$\lambda$ is readily obtained upon setting up the normality
condition of the above problem, namely,
 \beq
  \frac{\partial z}
  {\partial\lambda}\equiv
  \frac{1}{2n}{\rm tr}
  \left(\frac{\partial (\overline{\negr J}~\overline{\negr J}^T)}
  {\partial\lambda}
  \right) -
  \frac{1}{n}{\rm tr}\left(\negr K
  \frac{\partial\overline{\negr J}^T}{\partial\lambda}\right)=0
  \label{e:normal-1}
 \eeq
where we have used the linearity property of the trace and the
derivative operators. We calculate below the quantities involved:
 \bed
  \overline{\negr J}\overline{\negr J}^T=
  \left[
    \begin{array}{cc}
     n  &
     \lambda \sum_1^n \negr r_j^T\negr E^T \\
     \lambda \sum_1n3 \negr E\negr r_j &
     \lambda^2\sum_1^n \negr E\negr r_j
     \negr r_j^T\negr E^T
   \end{array}
  \right]
  {\rm , }
  \negr K\overline{\negr J}^T=
  \left[
    \begin{array}{cc}
     n &
     \lambda \sum_1^n \negr r_j^T\negr E^T \\
     \sum_1^n\negr E\negr k_j &
     \lambda\sum_1^n \negr E\negr k_j\negr r_j^T
     \negr E^T
   \end{array}
  \right]
 \eed
Thus,
 \bed
  \frac{\partial (\overline{\negr J}\overline{\negr J}^T)}
  {\partial\lambda}
  = \left[
  \begin{array}{cc}
  0 & \sum_1^n \negr r_j^T\negr E^T \\
  \sum_1^n \negr E \negr r_j &
  2\lambda\sum_1^n\negr E\negr r_j \negr r_j^T\negr E^T
  \end{array}
  \right]
  {\rm ~,~}
  \negr K\frac{\partial\overline{\negr J}^T}
  {\partial\lambda}
  = \left[
  \begin{array}{cc}
  0 &
  \sum_1^n \negr r_j^T\negr E^T \\
  \negr 0&
  \sum_1^n \negr E\negr k_j \negr r_j^T\negr E^T
  \end{array}
  \right]
 \eed
whence the normality condition (\ref{e:normal-1}) becomes
 \beqa
  \lambda \sum_1^n\|\negr r_j\|^2 - \sum_1^n \negr k_j^T\negr r_j=0
  \nonumber
 \eeqa
Now, if we notice that $\|\negr r_j\|$ is the distance $d_j$ from
the operation point $P$ to the center of the $j$th revolute, the
first summation of the above equation yields $n d_{\rm rms}^2$,
with $d_{\rm rms}$ denoting the root-mean-square value of the set
of distances $\set{d_j}1 n$, and hence,
 \beq
  \lambda = \frac{\sum_1^n \negr k_i^T\negr r_j}
  {nd_{\rm rms}^2}\quad\Rightarrow\quad
  l_{\mathcal P}=\frac{nd_{\rm rms}^2}{\sum_1^n \negr k_j^T\negr r_j}
  \label{e:l_P}
 \eeq
Thus, the conditioning length is defined so that
$\sqrt{n}\,d_{rms}$ is the geometric mean between $l_{\mathcal P}$
and the sum of the projections of the set $\set{\negr r_j}1n$ onto
the corresponding vectors of the set $\set{\negr k_j}1n$, as
illustrated in Fig.~\ref{figure:interpretation}.
\subsection{A Rotation of the Isotropic Set as a Rigid Body}
Since a rigid-body rotation of a set of isotropic points preserves
isotropy, we can find the orientation of this set, as
parameterized by the angle of rotation $\alpha$, that renders $z$
a minimum, for a given manipulator posture. Let this rotation be
$\negr R(\alpha)$, which can be expressed as \cite{Bottema:79}
 \beq
  \negr R(\alpha) = (\cos\alpha)\negr 1 + (\sin\alpha)\negr E
  \label{e:R(alpha)}
  \eeq
with \negr E defined in eq.(\ref{e:E}). Thus, upon rotating the
set $\mathcal K$ through an angle $\alpha$ about its centroid $C$
or, equivalently, about the operation point $P$, the isotropic
matrix \negr K becomes $\tilde{\negr K}$, and is given by
 \bseq
 \beq
  \tilde{\negr K}=
  \left[
  \begin{array}{cccc}
  1 & 1 & \cdots & 1 \\
  \negr E\negr R(\alpha) \negr k_1 &
  \negr E\negr R(\alpha) \negr k_2 &
  \cdots &
  \negr E\negr R(\alpha) \negr k_n
  \end{array}
  \right]
 \eeq
The objective function $z$ then becomes
 \beq
  z= \frac{1}{2n}
  \left[
  \tr(\overline{\negr J}~\overline{\negr J}^T)
  -2 \tr(\overline{\negr J} \tilde{\negr K}^T)
  + \tr(\tilde{\negr K}\tilde{\negr K}^T)
  \right]
  \quad \rightarrow \quad
  \min_\alpha
  \label{equation:min_angle}
 \eeq
 \eseq
Before setting up the normality conditions for the problem at
hand, we note that
 \beq
  \tilde{\negr K}\tilde{\negr K}^T =
    \left[
    \begin{array}{cc}
     n &
     (\sum_1^n \nigr ki)^T\negr E^T\negr R^T(\alpha) \\
     \negr R(\alpha)\negr E\sum_1^n \nigr ki&
     \negr R(\alpha)\negr E(\sum_1^n \nigr ki\nigr ki^T)
     \negr E^T\negr R^T(\alpha)
    \end{array}
    \right]
 \eeq
which can be shown to reduce to
 \beqa
  \tilde{\negr K}\tilde{\negr K}^T =
    \left[
    \begin{array}{cc}
      n &
      (\sum_1^n \nigr ki)^T\negr E^T\negr R^T(\alpha) \\
      \negr R(\alpha)\negr E\sum_1^n \nigr ki&
      k^2\nigr 1{2\times 2}
    \end{array}
    \right]
    \nonumber
 \eeqa
and hence,
 \beqa
    \frac{\partial[\tr(\tilde{\negr K}\tilde{\negr K}^T)]}
    {\partial\alpha}=0
    \nonumber
 \eeqa
On the other hand, $\overline{\negr J}$ is independent of
$\alpha$, and hence, the normality condition of problem
(\ref{equation:min_angle}) reduces to
 \beq
  \frac{\partial z}{\partial \alpha} \equiv -\frac{1}{n}
  \left[
  \tr\left(\overline{\negr J} \frac{\partial \tilde{\negr K}^T}
  {\partial \alpha}
  \right)
  \right]=0\label{e:normal-alpha}
 \eeq
We calculate below the partial derivative required above:
 \beqa
  \frac{\partial \tilde{\negr K}}{\partial \alpha}=
    \left[
    \begin{array}{cccc}
      0 & 0 & \cdots & 0 \\
     \negr R^\prime(\alpha)\negr E\nigr k1&
     \negr R^\prime(\alpha)\negr E\nigr k2&\cdots&
     \negr R^\prime(\alpha)\negr E\nigr kn
    \end{array}
    \right]
  \nonumber
 \eeqa
where $\negr R^\prime(\alpha)$ can be expressed, in light of
relation (\ref{e:R(alpha)}), as
 \beqa
  \negr R^\prime(\alpha)=\negr E\negr R(\alpha)
  \nonumber
 \eeqa
and hence,
 \beqa
  \frac{\partial \tilde{\negr K}}{\partial \alpha}=
    \left[
    \begin{array}{cccc}
      0&0&\cdots&0 \\
     \negr E\negr R(\alpha)\negr E \nigr k1&
     \negr E\negr R(\alpha)\negr E \nigr k2&\cdots&
     \negr E\negr R(\alpha)\negr E \nigr kn
    \end{array}
    \right]
  \nonumber
 \eeqa
whence,
 \beqa
  \tr\left(\overline{\negr J} \frac{\partial \tilde{\negr K}^T}
  {\partial \alpha}
  \right)&=& \lambda\tr\{\sum_1^n \negr E\nigr rj[\negr E\negr R(\alpha)
  \negr E\nigr kj]^T\} =\lambda \sum_1^n \negr r_j^T\negr E^T\negr E
  \negr R(\alpha)\negr E\nigr kj\nonumber\\
  &=&\lambda \sum_1^n \negr r_j^T\negr R(\alpha)\negr E \nigr kj
  \nonumber
  \eeqa
Now, under the plausible assumption that $l_{\mathcal P}$ is
finite, $\lambda\ne 0$, and hence, the normality condition
(\ref{e:normal-alpha}) reduces to
 \beqa
  \sum_1^n \negr r_j^T\negr R(\alpha)\negr E \nigr kj=0
  \nonumber
 \eeqa
Further, substitution of expression (\ref{e:R(alpha)}) into the
above expression leads to
 \beq
  (\cos\alpha)\sum_1^n \negr r_j^T\negr E \nigr kj
  -(\sin\alpha)\sum_1^n \negr r_j^T \nigr kj=0\label{e:normal-alpha-alt}
 \eeq
Therefore, the value of $\alpha$ minimizing the distance of
$\overline{\negr J}$ to $\tilde{\negr K}(\alpha)$ is, for the
given posture $\mathcal P$,
 \beq
  \alpha=\arctan\left[\frac{\sum_1^n \negr r_j^T\negr E \nigr kj}
  {\sum_1^n \negr r_j^T \nigr kj}\right]
  \equiv
  \arctan\left[\frac{(1/n)\sum_1^n \negr r_j^T\negr E \nigr kj}
  {(1/n)\sum_1^n \negr r_j^T \nigr kj}\right]
  \label{e:alpha-opt}
 \eeq
Thus, the angle $\alpha$ through which the given isotropic set
$\mathcal K$ is to be rotated in order to obtain the conditioning
length of the manipulator pose $\mathcal P$ is given as the
$\arctan$ function of the ratio of a numerator $N$ to a
denominator $D$, whose geometric interpretations are
straightforward: $D$ is simply the mean value of the projections
of the $\nigr rj$ vectors onto their corresponding $\nigr kj$
vectors. Now, since $\negr E\nigr kj$ is vector $\nigr kj$ rotated
$90^\circ$ counterclockwise, $N$ is the mean value of the
projections of the $\nigr rj$ vectors onto their corresponding
$\negr E\nigr kj$ vectors. We can call the latter the {\em
transverse projections} of the said vectors. Once we have found
the optimum value $\alpha_{\rm opt}$ of $\alpha$ for a given
manipulator posture, we redefine, for conciseness,
 \beq
  \negr K\quad\leftarrow\quad\tilde{\negr K}(\alpha_{\rm opt})
  \label{e:K-redef}
 \eeq
With $\alpha_{\rm opt}$ known, $l_{\mathcal P}$ is readily
computed from eq.(\ref{e:l_P}). Now, if we regard the columns of
$\overline{\negr J}$ and \negr K as 3-dimensional vectors, then a
rotation \negr Q of the corresponding 3-dimensional space about an
axis normal to the plane of the sets $\set{\nigr rj}1n$ and
$\set{\nigr kj}1n$ through an angle $q$ can be represented as
 \beqa
  \negr Q=
   \left[\begin{array}{cccc}
     1 &
     \negr 0^T \\
     \negr 0 &
     \negr R(q)
   \end{array}
   \right]
   \nonumber
 \eeqa
where $\negr R(q)$ is a $2\times 2$ rotation matrix similar to
$\negr R(\alpha)$, as defined in eq.(\ref{e:R(alpha)}). Hence,
under rotation \negr Q, $\overline{\negr J}$ and \negr K change as
described below: \beqa
 \negr Q\overline{\negr J}&=&
 \left[\begin{array}{cc}
    1   & \negr 0^T \\
    \negr 0 & \negr R(q)
 \end{array}
 \right]
 \left[\begin{array}{cccc}
   1 & 1 & \cdots & 1 \\
   (1/ l_{\mathcal P})~ \negr E \negr r_1 &
   (1/ l_{\mathcal P})~ \negr E \negr r_2 &
   \cdots &
   (1/ l_{\mathcal P})~ \negr E \negr r_n
 \end{array}
 \right]
 \nonumber\\
 &=&
 \left[\begin{array}{cccc}
         1     &   1     &  \cdots  &     1 \\
         \negr R(q)\negr E\nigr r1&\negr R(q)\negr E\nigr r2&
         \cdots & \negr R(q)\negr E \nigr rn
 \end{array}
 \right]
 \nonumber\\
 \negr Q\negr K&=&
 \left[\begin{array}{cccc}
         1     &   1     &  \cdots  &     1 \\
         \negr R(q)\negr E\nigr k1&\negr R(q)\negr E\nigr k2&
         \cdots & \negr R(q)\negr E \nigr kn
 \end{array}
 \right]
 \nonumber
 \eeqa
Now it is apparent that $z$, as defined in
eq.(\ref{e:least-squares}), is invariant under a rotation $\negr
R(q)$ of the sets $\mathcal S$ and $\mathcal K$. Indeed, under
such a rotation,
 \beqa
  z_{\negr Q}& \equiv &\frac{1}{2}\frac{1}{n}\tr (
  \negr Q\overline{\negr J}\overline{\negr J}^T\negr Q^T
  -\negr Q\negr K\overline{\negr J}^T\negr Q^T
  -\negr Q\overline{\negr J}\negr K^T\negr Q^T
  +\negr Q\negr K\negr K^T\negr Q^T)\nonumber\\
  &=&\frac{1}{2}\frac{1}{n}\tr [
  \negr Q(\overline{\negr J}\overline{\negr J}^T
  -\negr K\overline{\negr J}^T - \overline{\negr J}\negr K^T
  +\negr K\negr K^T)\negr Q^T]\nonumber
 \eeqa
If we recall relation (\ref{e:product-trace}), the above
expression becomes \beqa z_{\negr Q}& \equiv
&\frac{1}{2}\frac{1}{n}\tr [\negr Q^T\negr Q(\overline{\negr
J}\overline{\negr J}^T -\negr K\overline{\negr J}^T -
\overline{\negr J}\negr K^T +\negr K\negr K^T)]=z \eeqa We thus
have proven
\begin{Lem}
The distance of $\overline{\negr J}$ to \negr K is invariant under
a rotation of the sets $\mathcal S$ and $\mathcal K$.
\label{l:dist-inv}
\end{Lem}
\subsection{The Optimum Posture}\label{ss:optimum-posture}
It is now apparent that we can always orient the $\mathcal K$ set
optimally, so that, for any posture $\mathcal P$, $\overline{\negr
J}$ lies a minimum distance from the corresponding matrix \negr K.
Moreover, by virtue of Lemma~\ref{l:dist-inv}, a rotation of the
whole manipulator as a rigid body about its first joint, i.e., a
motion of the manipulator with all its joints but the first one
locked, does not affect $z$. That is, $z$ is a function of only
$\set{\theta_i}2n$, which can thus be termed the set of {\em
conditioning-joint variables}, the associated joints being the
{\em conditioning joints}. Now we aim at finding the {\em optimum
posture} ${\mathcal P}_o$ that yields a dimensionless Jacobian
$\overline{\negr J}$ lying a minimum distance to the reference
matrix \negr K. To this end, we adopt a given set $\mathcal K$ at
a given orientation at the outset, which thus leads to a constant
matrix \negr K in the process of finding ${\mathcal P}_o$, the
optimum orientation of $\mathcal K$ being readily determined from
eq.(\ref{e:alpha-opt}) once ${\mathcal P}_o$ has been found. Thus,
in the derivations that follow, $\theta_1$ can be set arbitrarily
equal to zero, or to any other constant value, for that matter. We
now aim at solving the problem \beq z=\frac{1}{2} \frac{1}{n}
\tr(\overline{\negr J}\overline{\negr J}^T -2\negr
K\overline{\negr J}^T+\negr K\negr K^T) \quad \rightarrow \quad
\min_{\{\theta_i\}_2^n}
        \label{e:optimize-P}
\eeq An attempt to solving this problem using the approach of the
foregoing sections proved to be impractical. Indeed, since $z$
depends on the set of conditioning joint variables both via the
set $\set{\nigr rj}1n$ and via $\lambda$, this dependence leads to
a normality condition that does not lend itself to a closed-form
solution. As a consequence, the said normality condition does not
lead to a direct geometric interpretation of the optimum posture.

We thus follow a different approach here. For each posture, the
value of $\lambda$ is computed using eq.(\ref{e:l_P}), while
$\alpha$ is computed with the eq.(\ref{e:alpha-opt}). With the
foregoing expressions substituted into the expression of $z$ given
in eq.(\ref{e:optimize-P}), the corresponding normality conditions
for angles ${\theta_i}^n_2$ yield a system of algebraic equations
in the foregoing conditioning variables that are amenable to
solutions using modern methods, like polynomial continuation,
Gr\"obner bases, or resultant methods \cite{Nielsen:98}, that
yield all roots of the problem at hand. These roots then lead to
the globally-optimum posture ${\mathcal P}_o$.
\subsection{The Characteristic Length}\label{ss:char-length}
The optimum postures of a given manipulator, i.e., those with a
Jacobian matrix closest to a corresponding model matrix \negr K
are thus found upon solving the optimization problem
(\ref{e:optimize-P}). Moreover, the conditioning length associated
with the posture yielding a global minimum of the foregoing
distance is defined as the {\em characteristic length} of the
manipulator at hand. Prior to discussing some examples, we would
like to find out whether the characteristic length thus found
bears a minimality geometric property, e.g., whether the
characteristic length is the minimum conditioning length of the
manipulator over its whole workspace. To this end, we rewrite the
objective function $z$ in the form \beq z=\frac{1}{2n}(\lambda^2 n
d_{\rm rms}^2 - 2\lambda\sum_1^n \negr k_j^T \nigr rj + \sum_1^n
\|\nigr kj\|^2)\label{e:z-alt} \eeq Upon substitution of the sum
$\sum_1^n \negr k_j^T \nigr rj$ in terms of the optimum value of
$\lambda$ found in eq.(\ref{e:l_P}) into eq.(\ref{e:z-alt}), we
obtain \beq z=\frac{1}{2n}\sum_1^n \|\nigr kj\|^2 -
\frac{1}{2}\lambda^2 d_{\rm rms}^2\equiv \frac{1}{2n}\sum_1^n
\|\nigr kj\|^2 -\frac{1}{2}\left(\frac{d_{\rm rms}}{l_{\mathcal
P}}\right)^2 \label{e:z-l_P} \eeq It is apparent from the above
expression that minimizing $z$ is not equivalent to minimizing
$l_{\mathcal P}$, but rather to maximizing the ratio $d_{\rm
rms}/l_{\mathcal P}$. Hence, minimizing $z$ is equivalent to
minimizing the inverse ratio, i.e., $l_{\mathcal P}/d_{\rm rms}$.
In other words, minimizing $z$ is equivalent to minimizing the
ratio of the conditioning length to the rms value of the distances
of the joint centers from the operation point.
\subsection{Examples: A Three-DOF Planar Manipulator}
\subsubsection{An Isotropic Manipulator}
\begin{figure}[!hbt]
\begin{center}
  \begin{tabular}{cc}
  \begin{minipage}[t]{60 mm}
  \begin{center}
  \includegraphics[width= 45mm,height= 40mm]{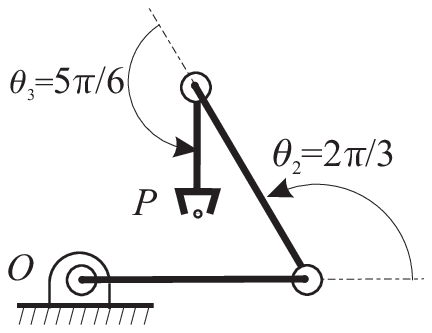}
  \caption{An isotropic posture with $\theta_1=0^\circ$}
  \protect\label{figure:isotropic_configuration}
  \end{center}
  \end{minipage} &
  \begin{minipage}[t]{90 mm}
  \includegraphics[width= 90mm,height= 70mm]{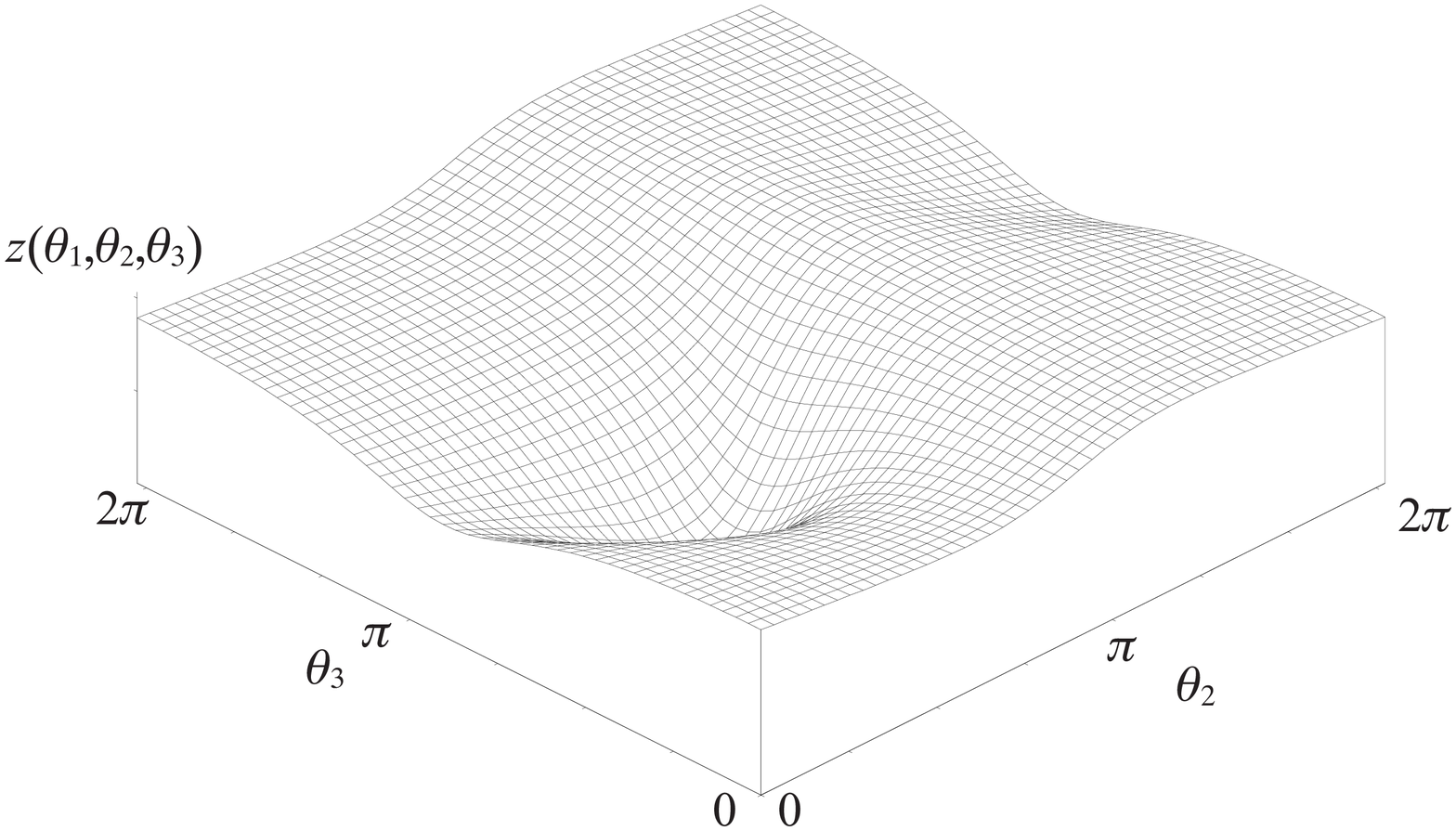}
  \caption{Objective function of the isotropic manipulator}
  \protect\label{figure:fonction_z_isotrope}
  \end{minipage}
  \end{tabular}
  \end{center}
\end{figure}

In the first example, we have $a_1= a_2 = l$ and $a_3= \sqrt{3}
\,l/3$, with the $\set{\nigr rj}13$ vectors given by
 \beqa
 \negr r_1= l
 \left[\begin{array}{r}
   \cos(\theta_1) + \cos(\theta_{12}) + \cos(\theta_{123})\\
   \sin(\theta_1) + \sin(\theta_{12}) + \sin(\theta_{123})
 \end{array}
 \right] {\rm ~,~} \nonumber \\
 \negr r_2= l \left[\begin{array}{c}
  \cos(\theta_{12}) + \cos(\theta_{123})\\
  \sin(\theta_{12}) + \sin(\theta_{123})
 \end{array}
 \right] {\rm ~,~}
 \negr r_3=
 \frac{\sqrt{3} l}{3}
 \left[\begin{array}{c}
   \cos(\theta_{123})\\
   \sin(\theta_{123})
 \end{array}
 \right] \nonumber
 \eeqa
with the definition $\theta_{ij\ldots p}\equiv \theta_i +\theta_j
+ \ldots + p$. Moreover, the model matrix \negr K used for this
case is that found for the case of an isotropic set of three
points, as discussed in Subsection~\ref{ss:example-1}, and
reproduced below for quick reference: \beq \negr K=
\left[\begin{array}{ccc} 1 & 1 & 1 \\ -\sqrt{2} / 2 & -\sqrt{2} /
2 & ~\sqrt{2} \\
 ~\sqrt{6} / 2 &
-\sqrt{6} / 2 & 0
\end{array}
\right] \eeq One optimum posture found with the procedure
discussed in Subsection~\ref{ss:optimum-posture} is displayed in
Fig.~\ref{figure:isotropic_configuration}, the objective function
attaining a minimum of zero at this posture, which means that the
manipulator can match exactly an isotropic model matrix \negr K
within its workspace, the manipulator thus being termed {\em
isotropic}. The objective function attains the values displayed in
Fig.~\ref{figure:fonction_z_isotrope} over its whole workspace.

At the optimum posture, we have the values of the joint variables
given below:
 \beqa
  \theta_1=0^\circ,\quad\theta_2=120^\circ,\quad\theta_3=150^\circ
  \nonumber
 \eeqa
the conditioning length $l_{\mathcal P}$ being equal to
$(\sqrt{6}/6)\, l $, which is thus the characteristic length of
this manipulator, as found using an alternative approach in
\cite{Angeles:97}. Moreover, the normalized Jacobian
$\overline{\negr J}$ becomes
 \bed
  \overline{\negr J}=
  \left[\begin{array}{ccc}
   1&1&1\\
   -\sqrt{2}/2&-\sqrt{2}/2&\sqrt{2}\\
   \sqrt{6}/2&-\sqrt{6}/2&0
  \end{array}
  \right]
 \eed
\subsubsection{An Equilateral Manipulator}
In the second example, we assume that all the link lengths are
equal to $l$, and hence,
 \beqa
 \negr r_1&=& l \left[\begin{array}{r}
 \cos(\theta_1) + \cos(\theta_{12}) + \cos(\theta_{123})\\
 \sin(\theta_1) + \sin(\theta_{12}) + \sin(\theta_{123})
 \end{array}
 \right]
 {\rm ~, ~}
 \negr r_2= l \left[\begin{array}{c}
 \cos(\theta_{12}) + \cos(\theta_{123})\\ \sin(\theta_{12}) +
 \sin(\theta_{123})
 \end{array}
 \right]
 \nonumber \\
 \negr r_3&=& l \left[\begin{array}{c}
 \cos(\theta_{123})\\ \sin(\theta_{123})
 \end{array}
 \right]
 \eeqa
We call this manipulator {\em equilateral}.
\par
\begin{figure}[!hbt]
  \begin{center}
  \begin{tabular}{cc}
  \begin{minipage}[t]{60 mm}
  \begin{center}
  \includegraphics[width= 45mm,height= 40mm]{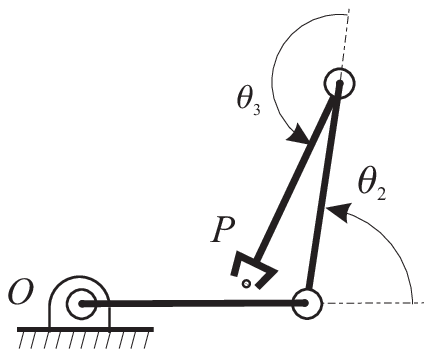}
  \caption{The posture closest to isotropy with $\theta_1=0^\circ$}
  \protect\label{figure:non_isotropic_configuration}
  \end{center}
  \end{minipage} &
  \begin{minipage}[t]{90 mm}
  \includegraphics[width= 90mm,height= 65mm]{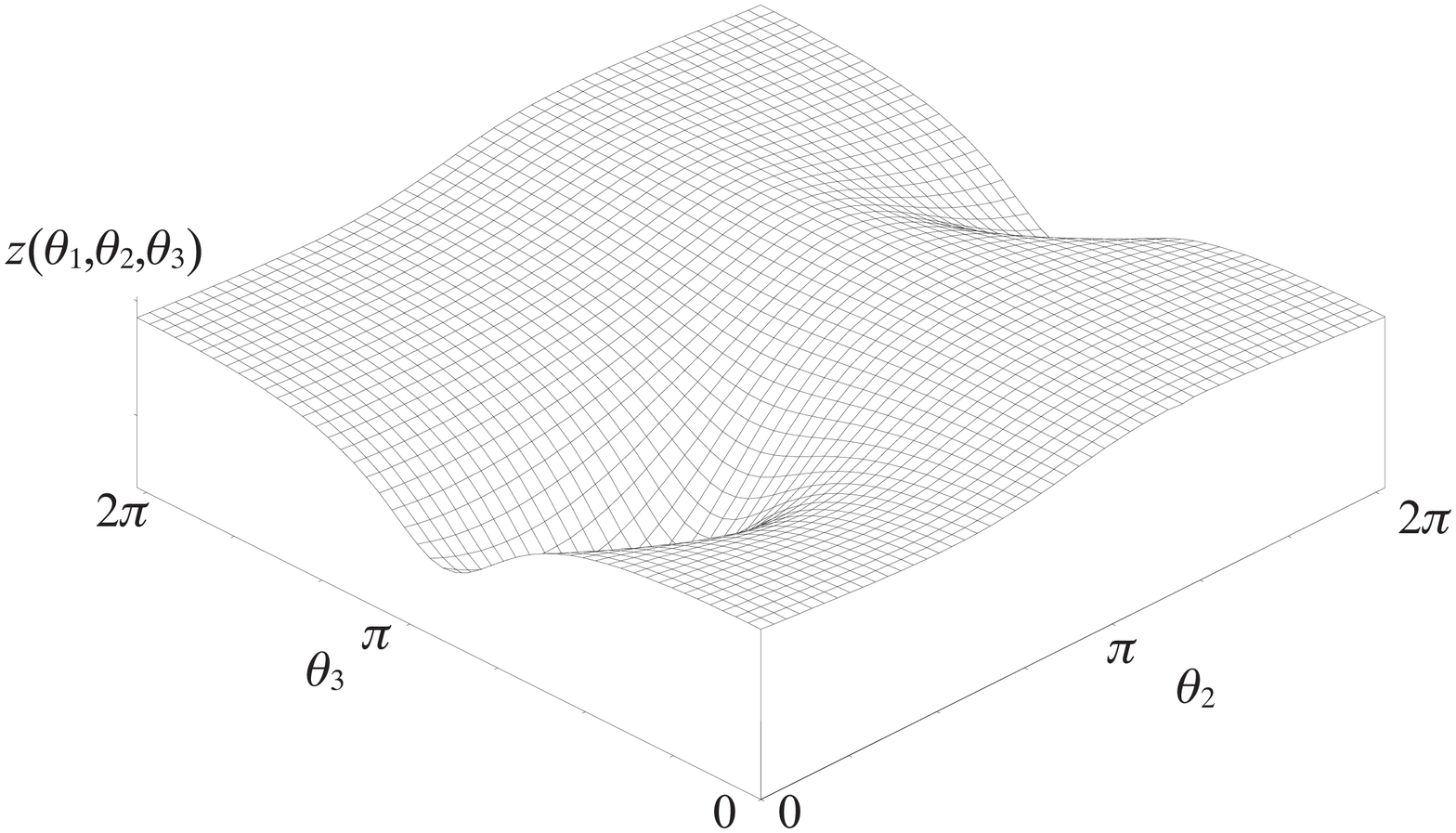}
  \caption{Objective function of the equilateral manipulator}
  \protect\label{figure:fonction_z_non_isotrope}
  \end{minipage}
  \end{tabular}
  \end{center}
\end{figure}

For this case, we use the same model matrix \negr K that we used
in the previous example, for the manipulator has the same number
of joints, and only one \negr K was found---up to a
reflection---for this number of joints. The minimization of the
objective function leads to the optimum values
 \bed
  \theta_1=0^\circ,\quad\theta_2= 81.8^\circ,\quad \theta_3= 155.2^\circ
 \eed
which correspond to a minimum value of $z=0.178$, the associated
characteristic length being $l_{\mathcal P}= 0.563\, l$. The
corresponding posture is displayed in
Fig.~\ref{figure:non_isotropic_configuration}, while the objective
function, evaluated throughout the workspace of the manipulator,
is displayed in Fig.~\ref{figure:fonction_z_non_isotrope}.

Finally, the normalized Jacobian at the posture of
Fig.~\ref{figure:non_isotropic_configuration} is
 \bed
  \overline{\negr J}=
  \left[
  \begin{array}{ccc}
  1 & 1 & 1\\
  -0.268 &  -0.268 & 1.489\\
  1.061 & - 0.714 & - 0.966
  \end{array}
  \right]
 \eed
\subsection{The Isocontours of the Objective Function}
\begin{figure}[!hbt]
\begin{center}
\begin{tabular}{cc}
\begin{minipage}[t]{80 mm}
\includegraphics[width= 70mm,height= 70mm]{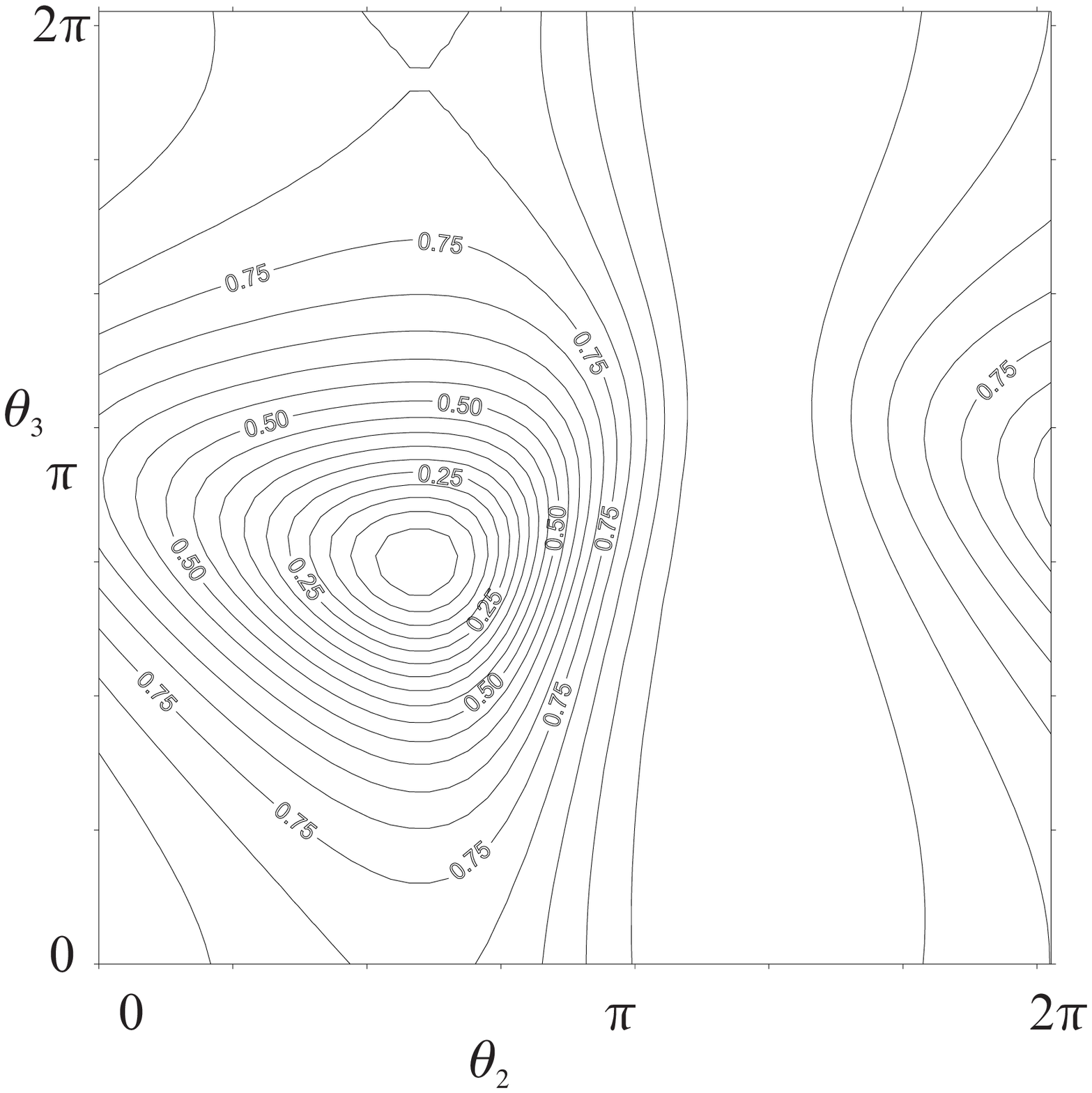}
\caption{The isocontours of the objective function of the
isotropic manipulator} \protect\label{figure:z_isotrope}
\end{minipage} &
\begin{minipage}[t]{80 mm}
\includegraphics[width= 70mm,height= 70mm]{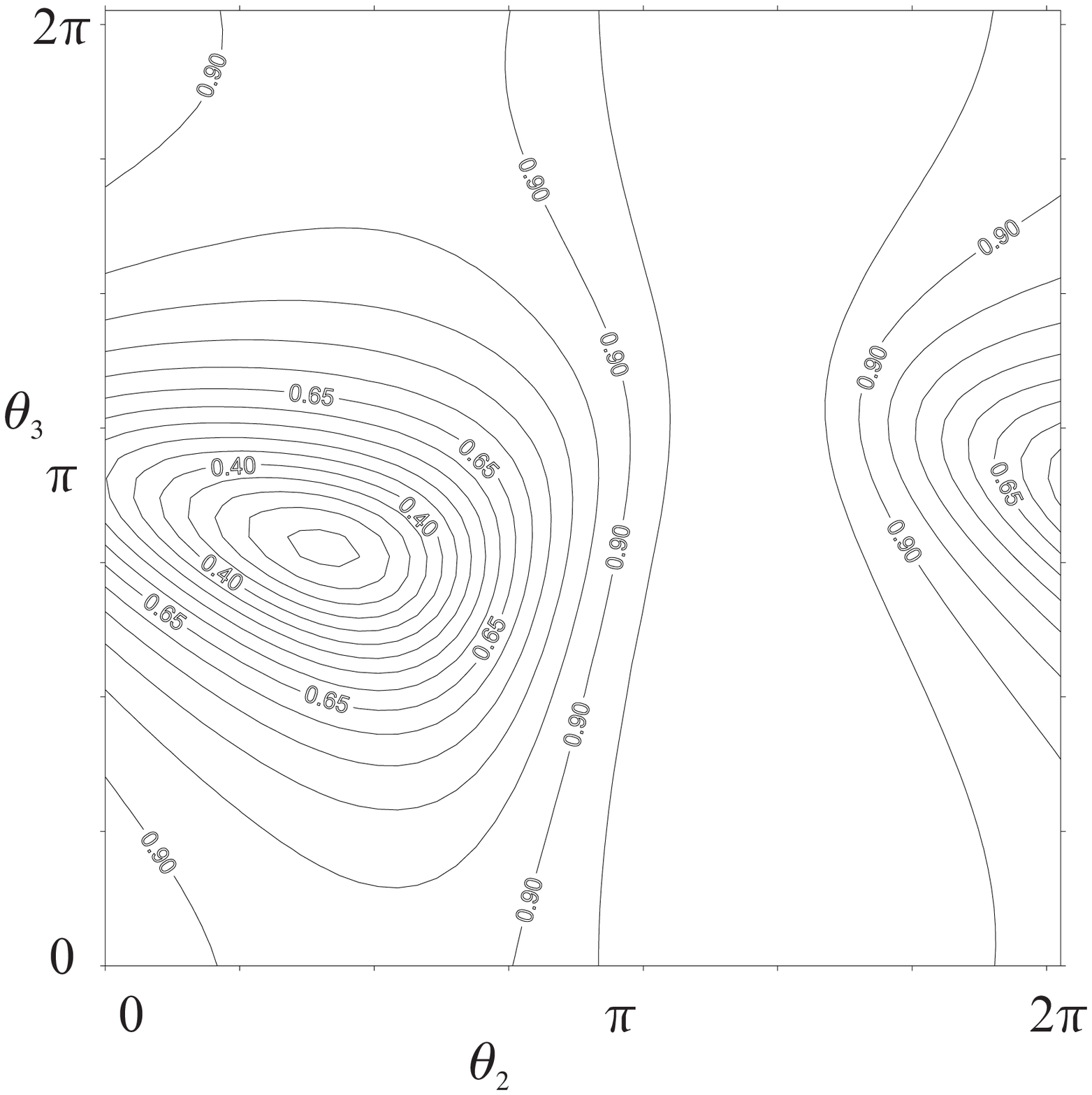}
\caption{The isocontours of the objective function of the
equilateral manipulator} \protect\label{figure:z_non_isotrope}
\end{minipage}
\end{tabular}
\end{center}
\end{figure}

The minimum of the objective function $z$ corresponds to the
posture closest to isotropy. At the other end of the spectrum, the
maximum of this function is attained at those singular postures
whereby the rank of the Jacobian matrix is two. The curves of
constant $z$-values, termed the {\em isocontours} of the
manipulator, can be used to define a performance index to compare
manipulators, as described below. The isocontours were obtained
with {\em Surfer}, a Surface Mapping System, for $\theta_2$ and
$\theta_3 \in [0,\,2\pi]$. The isocontours of the isotropic
manipulator, of the first example, are displayed in
Fig.~\ref{figure:z_isotrope}; those of the equilateral manipulator
in Fig.~\ref{figure:z_non_isotrope}.

\begin{figure}[!hbt]
  \begin{center}
  \begin{tabular}{cc}
  \begin{minipage}[t]{80 mm}
  \includegraphics[width= 70mm,height= 70mm]{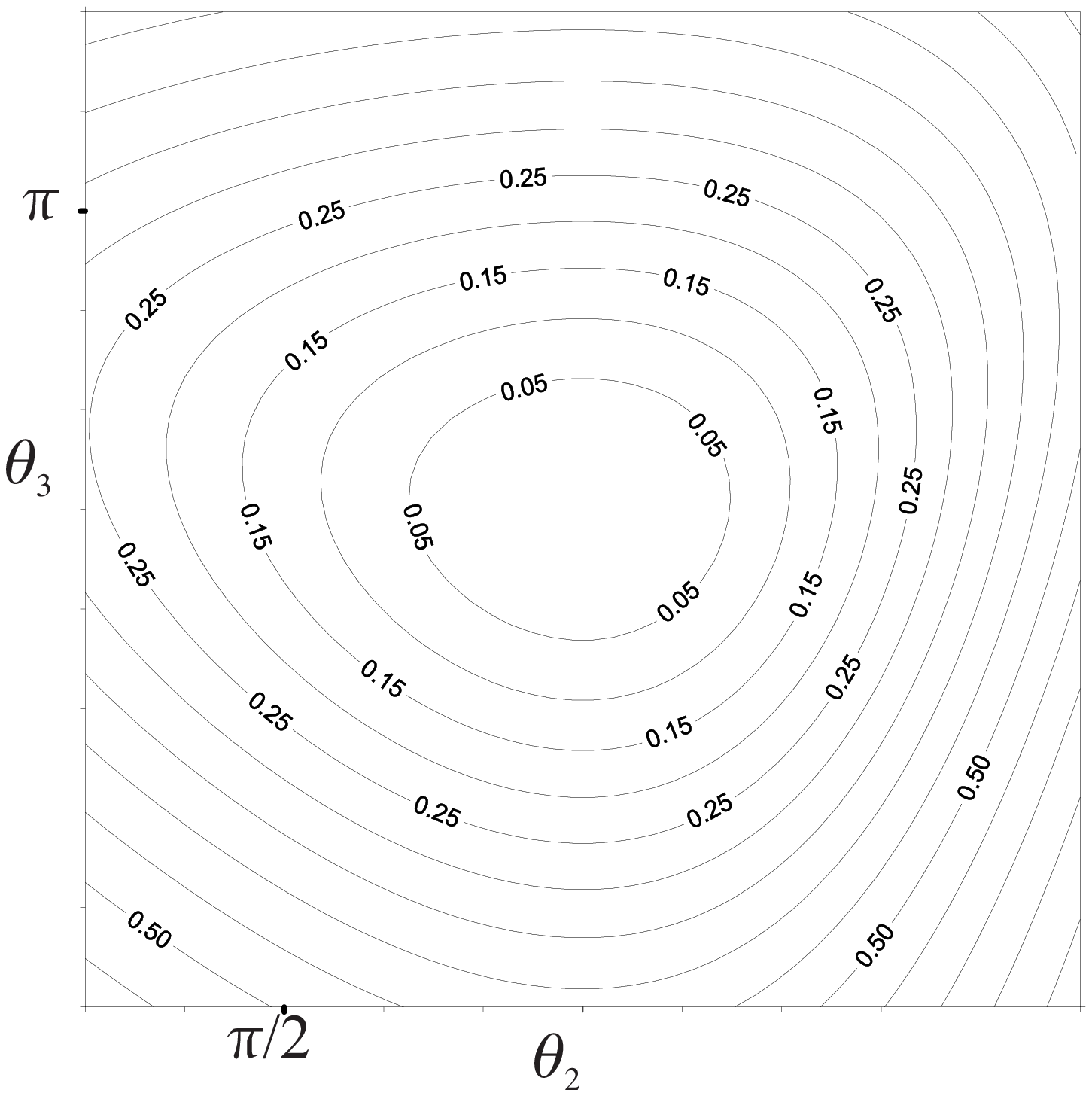}
  \caption{The isocontours closset to isotropy of the objective
  function of the isotropic manipulator}
  \protect\label{Figure:cercle}
  \end{minipage} &
  \begin{minipage}[t]{80 mm}
  \includegraphics[width= 70mm,height= 70mm]{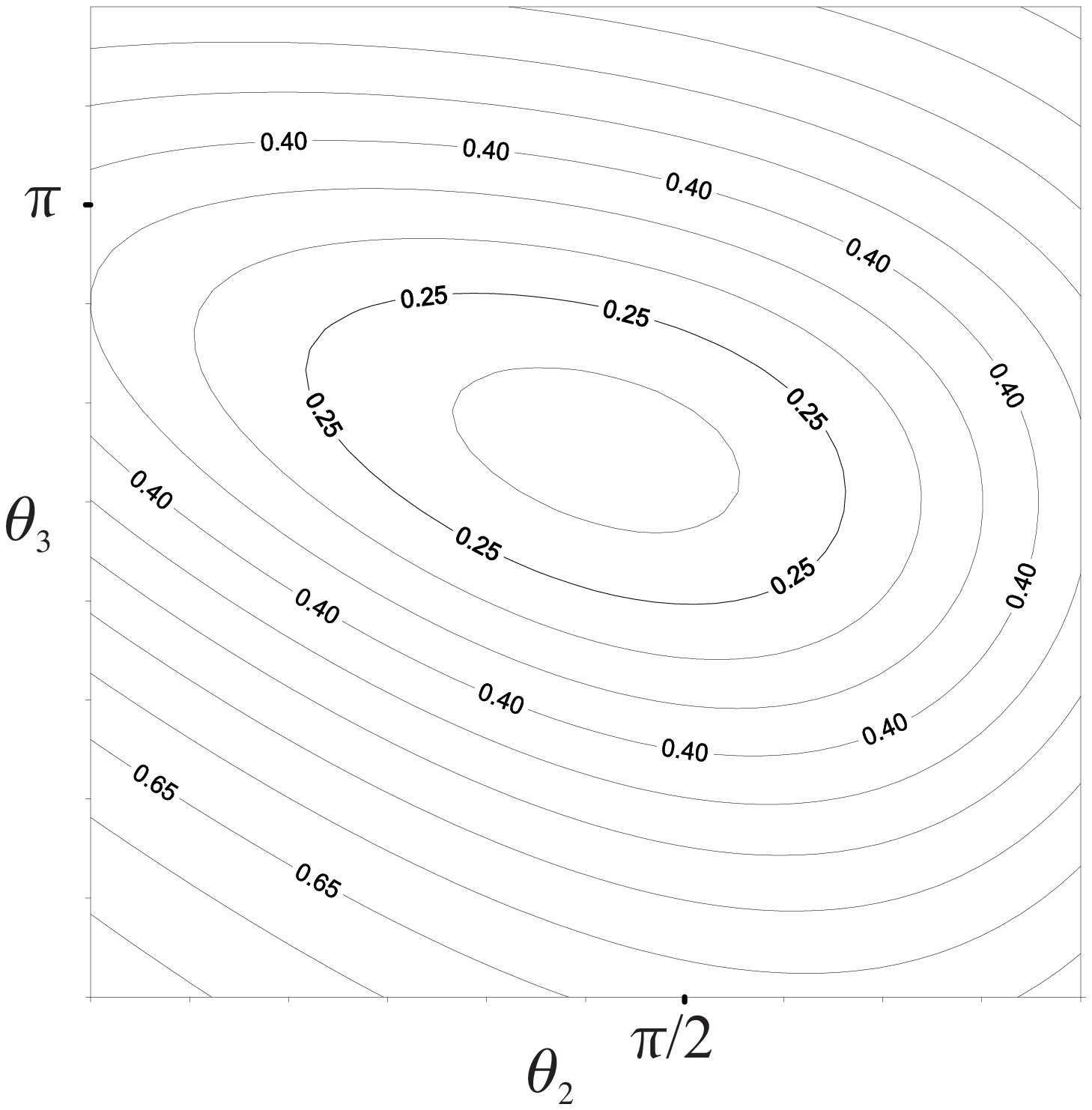}
  \caption{The isocontours closest to isotropy of the objective
  function of the equilateral manipulator}
  \protect\label{Figure:elipse}
  \end{minipage}
  \end{tabular}
  \end{center}
 \end{figure}

It is apparent from the two foregoing figures that the isocontours
can be closed or open. If closed, the curves enclose the optimum
point in the space of conditioning joints; in the second case, the
curves are periodic. The shape of the closed curves, additionally,
provides useful information on the manipulator performance: For
the isotropic manipulator, when the objective function is below
$0.25$, the curves are close to circular, as shown in
Fig.~\ref{Figure:cercle}; for the equilateral manipulator, these
curves are close to elliptical, as shown in
Fig.~\ref{Figure:elipse}. This means that, in the neighborhood of
an optimum, the isocontours behave in a way similar to the {\em
manipulability ellipsoid} \cite{Yoshikawa:85}: An isotropic
manipulator entails a manipulability ellipsoid with semiaxes of
identical lengths.
\section{Applications to Design and Control}
Manipulators are designed for a family of tasks, more so than for
a specific task---manipulator design for a specific task defeats
the purpose of using a manipulator, in the first place! The first
step in designing a manipulator, moreover, is to dimension its
links. It is apparent that from a purely geometric viewpoint, the
link lengths are not as important as the link-length ratios. Once
these ratios are optimally determined, the link lengths can be
obtained based on requirements such as maximum reach for a given
family of tasks, e.g., whether the manipulator is being used for
cleaning a wide-body or a regional aircraft. Now, the maximum
reach is directly proportional to the $d_{\rm rms}$ value of the
distance of the joint centers to the operation point at the
optimum posture, and hence, we can obtain the optimum link-length
ratios by assuming that $d_{\rm rms}$ is equal to one unit of
length. This means that minimizing the objective function $z$, as
given by eq.(\ref{e:z-l_P}), is equivalent to minimizing the {\em
normalized} conditioning length $l_{\mathcal P}$, where the
normalization is carried out upon dividing this length by $d_{\rm
rms}$. Furthermore, when deciding on the manipulator link-length
ratios, we may specify a certain {\em useful workspace region} as
a subset of the whole workspace. How to decide on the boundaries
of this region is something that can be done based on the value of
$z$, so that we can establish a maximum allowable value of $z$,
say $z_M < z_{\rm max}$, that we are willing to tolerate so as to
keep the manipulator far enough from singularities. In this
regard, the area enclosed by the isocontour $z=z_M$ will give a
nondimensional measure, and hence, a measure independent of the
scale of the manipulator, of the useful workspace region.

Under no constraints on the link-length ratios, for example, the
designer should choose the optimum ratios of the isotropic
manipulator of Fig.~\ref{figure:isotropic_configuration}. On the
other hand, when the manipulator is given, and it is desired to
control it so as to keep it away from singularities, function $z$
can be used again as a measure of the distance to singularities:
when $z$ attains its global maximum, the manipulator finds itself
at a rank-two singularity. A rank-one singularity occurs at a
local maximum. Furthermore, if a manipulator of given link-length
ratios---e.g., one out of a family of manipulators with identical
architectures, but of different scales, like the Puma 260, 560, or
760---is to be used for arc welding, then (a) the most suitable
dimensions should be chosen according with the dimensions of the
welding seam, and (b) the seam should be placed with respect to
the manipulator in such a way that, as the EE traces that seem
with the welding nozzle at a given angle with the seam, the
objective function $z$ must remain within a maximum value $z_M$.
This means that the seam should lie as close as possible to the
optimum posture of Fig.~\ref{figure:non_isotropic_configuration}.
Furthermore, note that a rotation of the manipulator about the
first joint axis, while keeping its other two joints locked, does
not perturb $z$, and hence, a set of optimum postures is
available. This set comprises the circle centered at the center of
the first joint, of radius $d_1$---the distance of the operation
point $P$ to the center of the first joint. This circle is similar
to the {\em isotropy circle} of isotropic manipulators
\cite{Angeles:97}, and thus, can be termed the {\em conditioning
circle}. Therefore, a good criterion to properly place the seam is
that the seam lie as close a spossible to the conditioning circle.

Finally, while detecting singularities of nonredundant robots is a
rather trivial task, detecting those of their redundant
counterparts is more involved, and a fast estimation of the
proximity of a given manipulator posture to singularity is always
advantageous. This estimation is provided by the objective
function $z$ proposed in this paper.

\section{Conclusions}
The conditioning length $l_{\mathcal P}$ was defined for a given
posture of a planar manipulator. This concept allows us to
normalize the Jacobian matrix so as to render it in nondimensional
form. We base the definition of the characteristic length on an
objective function $z$ that gives a geometric significance to the
conditioning length. Moreover, the objective function introduced
here is defined as a measure of the distance of the
normalized---nondimensional---Jacobian matrix to an isotropic
reference matrix. Isotropic sets of points in the plane are
defined as well as operations on these sets. The paper is limited
to planar manipulators, the treatment of spatial manipulators
being as yet to be reported.
\section*{Acknowledgements}
The first author acknowledges support from France's Institut
National de Recherche en Informatique et en Automatique. The
second author acknowledges support from the Natural Sciences and
Engineering Research Council, of Canada, and of Singapore's
Nanyang Technological University, where he completed the research
work reported here, while on sabbatical from McGill University.
\bibliographystyle{unsrt}

\section{Illustrations}

\end{document}